\def\year{2021}\relax
\documentclass[letterpaper]{article} 
\usepackage{aaai21}  
\usepackage{times}  
\usepackage{helvet} 
\usepackage{courier}  
\usepackage[hyphens]{url}  
\usepackage{graphicx} 
\usepackage{natbib}  
\usepackage{caption} 
\usepackage[switch]{lineno}
\usepackage{multirow}
\usepackage{booktabs}
\usepackage{tabularx}
\usepackage[ruled]{algorithm2e}
\usepackage{enumitem}
\setlist{nosep}

\title{Dynamically Grown Generative Adversarial Networks}
\author{
    Lanlan Liu\thanks{Work conducted during internship at Amazon.}, \textsuperscript{\rm 1}
    Yuting Zhang, \textsuperscript{\rm 2}
    Jia Deng, \textsuperscript{\rm 3}
    Stefano Soatto \textsuperscript{\rm 2}\\
}
\affiliations{
    \textsuperscript{\rm 1} University of Michigan, Ann Arbor\\
    \textsuperscript{\rm 2} Amazon Web Services \\
    \textsuperscript{\rm 3} Princeton University \\
    llanlan@umich.edu, yutingzh@amazon.com, jiadeng@cs.princeton.edu, soattos@amazon.com\\
}

\begin{document}
\maketitle
\begin{abstract}
Recent work introduced progressive network growing as a promising way to ease the training for large GANs, but the model design and architecture-growing strategy still remain under-explored and needs manual design for different image data. In this paper, we propose a method to dynamically grow a GAN during training, optimizing the network architecture and its parameters together with automation. The method embeds architecture search techniques as an interleaving step with gradient-based training to periodically seek the optimal architecture-growing strategy for the generator and discriminator. It enjoys the benefits of both eased training because of progressive growing and improved performance because of broader architecture design space. Experimental results demonstrate new state-of-the-art of image generation. Observations in the search procedure also provide constructive insights into the GAN model design such as generator-discriminator balance and convolutional layer choices.
\end{abstract}

\section{Introduction}

Generative Adversarial Networks (GANs)~\cite{goodfellow2014generative} have recently advanced in the literature of image generation and editing. However, training a GAN remains difficult for new data domains, especially for large-scale GANs. 
One of the main issues is that training a GAN can be unstable and difficult~\cite{arjovsky2017wasserstein,heusel2017gans}---as a two-player game, generator or discriminator could easily dominate the training and cause the training signal to vanish or explode. With bigger networks and higher resolutions, this issue gets worse with more complex network parameter space.

Recent work introduced the progressive growing of GANs~\cite{karras2018progressive} to ease the training for large GANs. It starts with generating low-resolution images with a small network and then progressively adds new layers to the network to generate higher-resolution images. As demonstrated by \cite{karras2018progressive}, it makes training large GANs easier and improve the quality of generated images in high resolutions.

However, in this progressive-growing process for architectures, the layers and growing schedules are predefined by researchers. There is a large space that still remains under-explored. For example, are symmetric generator and discriminator optimal? Are the layer choices optimal? 
In this paper, we propose an automatic method to explore this rich architecture space by dynamically growing a GAN during training, optimizing the growing strategy together with its network parameters.

Our Dynamically Grown GAN (DGGAN) method embeds architecture search techniques as an interleaving step with gradient-based training to periodically seek the optimal regarding the balancing between the generator and discriminator, 
choice of network units, and growing strategy. That is, we alternate between optimizing the generator and discriminator architecture and training the new architecture. 
More specifically, when optimizing the generator and discriminator architecture, our method grows layer(s) from previous architecture, with where to grow (generator or discriminator or both) and how to grow in the automatic framework. The new architecture is then trained with weight inheritance from the previous architecture.

Our method enjoys the benefits of both easing the optimization by progressively growing the architecture and exploring more architecture design space by architecture search. This combination is beneficial
to discover well-performing GANs, 
especially with high-resolution images. 
Compared to  progressively growing GANs~\cite{karras2018progressive} with a manually-designed growing strategy, our dynamic growing method explores much richer architecture space and growing strategies. More specifically, our method allows the generator and the discriminator to grow alone or together dynamically in the process, creating diverse and unconventional balance between them. 
Compared to prior work AutoGAN~\cite{gongautogan} and AGAN~\cite{wang2019agan}, which combine architecture search with GANs, we complement progressive growing with architecture search to eases the training of GANs with complicated architectures and high resolutions.
This novel perspective enables our method to work on higher resolutions while these prior works only worked on at most 48$\times$48.

Our experiments show that we achieve the new state-of-the-art on CIFAR-10 and the best performance among non part-based GANs on LSUN for image generation.
With further analysis of the thousands of models in our search procedure, we observe several practical conclusions on the GAN model design such as generator-discriminator  balance and convolutional layer choices.

Our main contributions are:
1) We propose a method to dynamically grow GANs, easing the training of GANs as well as exploring unconventional network architecture space;
2) We present the first automatic GAN that works on high-resolution images;
3) We provide constructive conclusions of generator and discriminator design choices using thousands of searched models;
4) We achieve new state-of-the-art image generation performance.

\section{Related Work}
\subsubsection{Improving GANs}
Since the proposal of GANs~\cite{goodfellow2014generative}, there have been several lines of work to improve GANs, including improving loss functions~\cite{arjovsky2017wasserstein,deshpande2018generative,berthelot2017began}, regularization techniques~\cite{miyato2018spectral,gulrajani2017improved,zhang2019consistency}, generation strategy~\cite{lin2019coco}, and architecture designs~\cite{li2019mad,nguyen2017dual,zhang2019self,radford2015unsupervised,karras2018progressive}.  Improved loss functions and regularization techniques are orthogonal to our work. 
Our work belongs to the architecture design line of work. While these works improve GAN training, they use manually-designed architectures. 

Progressive GAN (ProgGAN)~\cite{karras2018progressive} shows that progressively growing and training GAN from a smaller scale eases the training difficulties and improves the generation quality. Motivated by this observation, the search strategy in our DGGAN is in a progressive way. 
Despite this inspiration, our method is different from ProgGAN because ProgGAN is pre-designed and always grows both generator and discriminator symmetrically and simultaneously. Ours automatically searches how and what to grow, and generator and discriminator may not be symmetric. This difference allows us to explore a much richer architecture space than ProgGAN, resulting in better performance.

\subsubsection{Architecture Search with GANs}
AGAN~\cite{wang2019agan} and AutoGAN~\cite{gongautogan} explore architecture search with GANs, which are closely related to our work. 
AGAN~\cite{wang2019agan} learns one RNN controller with the REINFORCE algorithm to search for the architecture of both generator and discriminator simultaneously.
Our DGGAN differs from AGAN because our dynamic growing strategy is more flexible: it allows to grow G alone, or D alone, or both. Moreover, our method alternate between growing an architecture and training its weight while AGAN does not have such a training strategy.

AutoGAN~\cite{gongautogan} uses RNN controllers with REINFORCE to search for the generator architecture only. Their discriminator is not searched but constructed with a predefined strategy. It is stated that when searching for both they observe ``such two-way NAS will further deteriorate the original unstable GAN training, leading to highly oscillating training curves and often failure of convergence.'' Our DGGAN, however, deals with the difficulties in training unstable GANs and successfully searches for both generator and discriminator.
Beyond the methodology difference, both prior works demonstrate image generation with at most 48$\times$48 resolution. Our method, however, supports 256$\times$256.

\subsubsection{Neural Architecture Search}
Neural Architecture Search (NAS)~\cite{zoph2018learning,zoph2016neural} aims to reduce the human intervention by automating network design. 
Researchers use Reinforcement Learning~\cite{zoph2018learning,zoph2016neural,pham2018efficient}, Evolutionary Algorithm~\cite{real2019regularized}, gradient based~\cite{luo2018neural,liu2018darts}, Random Search~\cite{xie2019exploring} and progressive search~\cite{liu2018progressive}  for image classification task.
In addition, there are also recent works applying NAS on  segmentation~\cite{liu2019auto,chen2018searching}, machine translation~\cite{so2019evolved}, and transfer learning~\cite{wong2018transfer}. 
Our method differs from them because 1) we utilize the fact that GAN has two competing components, which does not exist in the tasks above, by allowing each component to grow alone or together; 2) we embed progressive training into architecture search, which is important 
to models with unstable training, such as GANs. As shown in AutoGAN~\cite{gongautogan}, a naive two-way NAS will deteriorate the original unstable GAN training and sometimes lead to failure of convergence.

One related work in this scope is progressive NAS~\cite{liu2018progressive}. 
It does progressive search inside a cell architecture and stacks the found cell to construct the full network for evaluation. Despite the similarity of wording choice, their progressive search is layer by layer inside a cell but our progressive growing involves growing from lower resolution to higher resolution. 
Also, we directly construct the whole network(s) while \cite{liu2018progressive} constructs a cell architecture and stack it for final architecture.
In addition, as a NAS method, it also has the differences described in the last paragraph.

\section{Dynamically Grown GAN}
\subsection{Dynamic Growing Overview}
\label{sec:method}

\begin{figure*}[!tb]
\begin{center}
\includegraphics[width=0.9\linewidth]{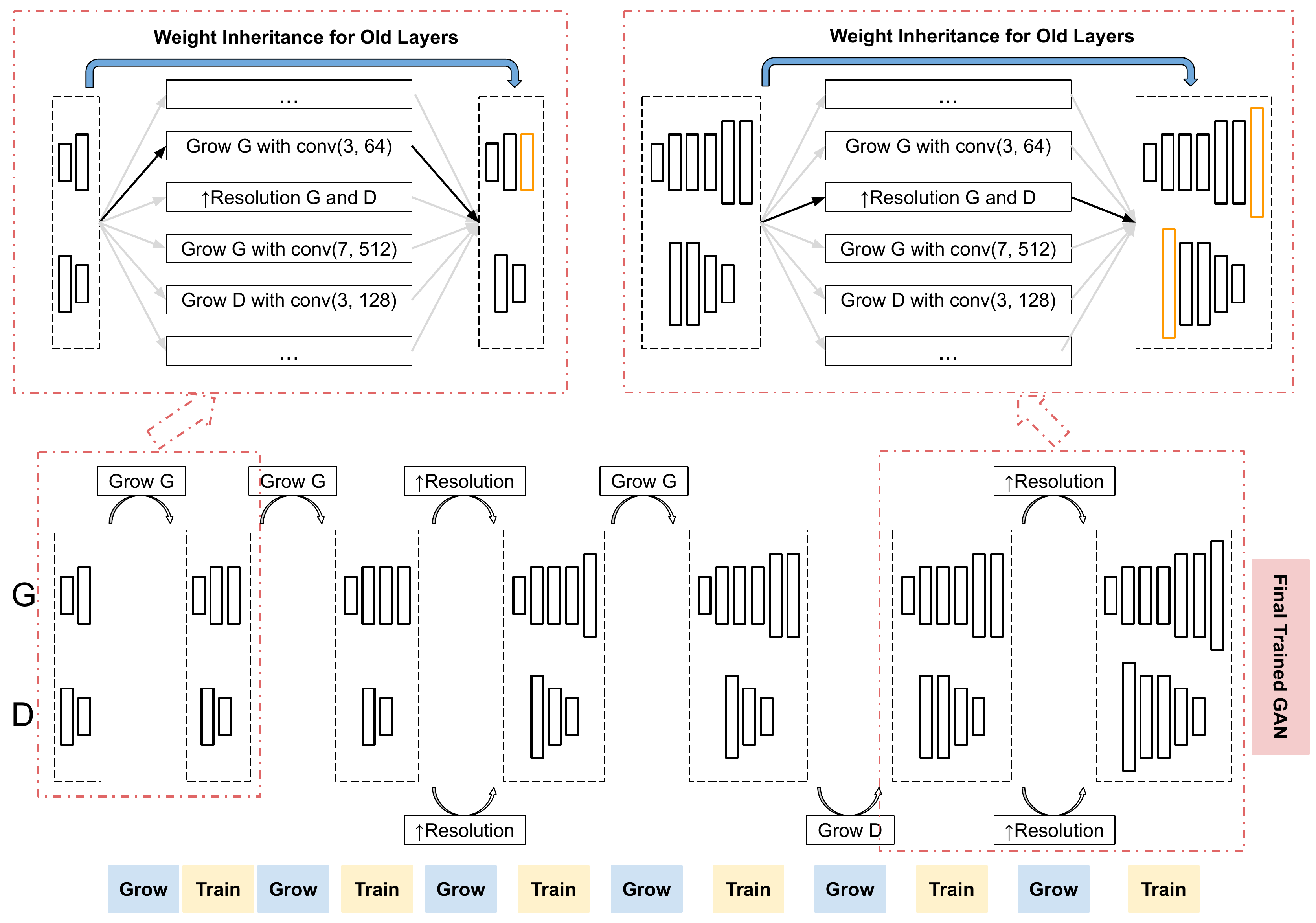}
\end{center}
  \caption{Overview of the Dynamically Grown GAN. Bottom: The dynamic growing process. We alternate between growing the architecture and training the weights of the new architectures. Each growing step chooses among actions including growing the generator (G) with a certain convolution layer, or growing discriminator (D) with a certain convolution layer, or growing both G and D to a higher resolution. In each training step, the new architectures inherit the weights from the old architectures. Top: Examples of growing steps.}
\label{fig:overview}
\end{figure*}

DGGAN embeds architecture search techniques with gradient-based training to periodically seek the optimal regarding network architecture and network weights. 
We alternate between the growing and training steps to grow a small-scale GAN to the full-scale GAN dynamically and automatically, as shown in Fig.~\ref{fig:overview}. We use the Wasserstein distance loss with gradient penalty~\cite{arjovsky2017wasserstein,gulrajani2017improved}. 

The growing at each step is dynamic instead of static as in the conventional progressive GAN method.
At the growing step, our method determines which actions in the search space to take, i.e., \emph{where} to grow \emph{what} kind of layers as in Fig.~\ref{fig:overview}(top). In contrast, in the conventional progressive GAN method, the growing schedule is fixed and manually-designed.
Given existing architectures, noted as parent architectures, we expand parent architectures to new architectures by the growing actions (e.g. add a layer with 256 filters and the filter size of 3). We note the new architectures as child architectures. 

At the training step, the child architectures are trained with weight inheritance from the parent architectures. More specifically, a new child architecture contains all layers in the parent architecture as well as the new layer(s) grown. We inherit the weights of the common layers from its parent architecture by initializing the weights of the child architecture with the trained weights of the parent architecture. The newly grown layer(s) that only exist in child candidates are initialized randomly.
This weight inheritance method provides two benefits: it eases the optimization difficulties for the child architectures. It also reduces the training time needed for each new candidate.

\subsubsection{Generator and Discriminator Base Architecture}
Following ProgGAN~\cite{karras2018progressive}, the base architectures of generator and discriminator are two small scale networks that operate on a low resolution with $d_0\times d_0$ pixels. The generator takes a randomly-sampled vector as input and outputs an image with resolution $d_0\times d_0$. The discriminator takes an image with  resolution $d_0\times d_0$ and outputs a single score. More specifically in our experiments, CIFAR-10 takes random vectors with size 128 and others take random vectors with size 512. $d_0$ is 8 in our experiments.
When growing a generator, we grow near the end of the network before the last convolution layer. When growing a discriminator, we grow near the beginning of the network after the first convolution layer. Growing heads are indicated as the orange layers in Fig.~\ref{fig:overview}(top). The structures of the base architectures thus can be maintained during growing. This choice is following prior work~\cite{karras2018progressive} as in their study, growing in other locations does not lead to better performance.
Note that our method allows us to grow the generator or discriminator alone or both. That is, they do not need to be grown symmetrically. 

\subsubsection{Action Search Space}
The action search space includes:
1) growing the generator with a convolution layer with filter sizes from \{3, 7\} and number of filters from \{32, 64, 128, 256, 512, 1024\}, with padding so that the resolution does not change;
2) growing the discriminator with a convolution layer with filter sizes from \{3, 7\} and number of filters from \{32, 64, 128, 256, 512, 1024\}, with padding so that the resolution does not change;
3) growing both generator and discriminator by adding a fade-in block~\cite{karras2018progressive} to double the resolution.
The fade-in block, introduced in ProgGAN~\cite{karras2018progressive},  can smoothly transit the network from lower resolution to higher resolution and avoid sudden dramatic changes in well-trained low-resolution networks. Sudden dramatic changes without the fade-in block may cause the training to diverge.  
More specifically, to transit from lower resolution to higher resolution, the fade-in block uses a weighted sum of both the lower resolution path and the higher resolution path. During training, we gradually reduce the weight on the lower resolution path from 1 to 0 and increases the weight of the higher resolution path from 0 to 1, so that it transits softly. 

This search space allows us to grow either or both of the generator and discriminator at each step.
We thus can explore various architecture combinations of the generator and discriminator without symmetric constraint while ensuring the image resolution is consistent between both.
An example of such asymmetric GAN is shown in Fig.~\ref{fig:overview}(bottom).

\begin{algorithm}[!b]
\SetAlgoLined
 $\mathcal{A}$ = \{actions\}\;
 $\mathcal{P}$ = \{initial candidates\}\;
 Train and evaluate each candidate in $\mathcal{P}$\;
 \While{not reaching maximum number of layers}{
   $\mathcal{C}$ = \{\}\;
   \For{$\normalfont{}\textbf{each}$ $(P_i,A_j)\in\mathcal{P}\times\mathcal{A}$}{
        Add $P_i + A_j$ into $\mathcal{C}$ with probability $p$;
    }
   Train and evaluate each candidate in $\mathcal{C}$\;
   $\mathcal{C}’$ = keep top-$K$ candidates of $\mathcal{C}’$\;
   $\mathcal{P}$ = $\mathcal{C}'$\;
}
\caption{Top-K Greedy Pruning Algorithm}
\label{algo:topk}
\end{algorithm}

\subsection{Search Algorithm}

In the growing steps discussed above, 
the number of potential architecture candidates will grow exponentially with respect to the search depth. Due to resource limits, it is infeasible to evaluate all of the candidates. We thus use random sampling and greedy pruning to reduce the number of candidates. 
More specifically, at each growing step, we reduce the number of parent candidates by greedy pruning and reduce the number of actions by random sampling. In particular, we keep the top $K$ parent candidates at each step and expand to child candidates only with these parents, similar to beam search. With the $T$ actions, we randomly sample actions with a probability $p$. The details are in Algorithm~\ref{algo:topk}. 
This pruning algorithm reduces the number of candidates from exponential to linear, with respect to search depth. 
Larger $p$ or $K$ leads to a better exploration of the search space but also greater computational cost. We will further discuss the computational efficiency in Sec. Analysis.

\section{Experimental Results}

We evaluate DGGAN against manually designed ProgGAN and other recent GAN models on CIFAR-10 and LSUN. 
We use the most popular PyTorch implementation\footnotemark of ProgGAN to obtain comprehensive ProgGAN results and to implement our DGGAN. 
Ablative analysis is on CIFAR-10.

\addtocounter{footnote}{-1}
\stepcounter{footnote}\footnotetext{\label{foot:1}\url{https://github.com/facebookresearch/pytorch_GAN_zoo}}
\subsection{Datasets and Evaluation Metrics}

CIFAR-10~\cite{krizhevsky2009learning} contains 50k 32$\times$32 training images. It is a small but effective testbed for GANs, including recent automatic GANs~\cite{gongautogan,wang2019agan}. 
LSUN~\cite{yu2015lsun} has over a million 256$\times$256 bedroom images for training. 
We use Frechet Inception Distance (FID)~\cite{heusel2017gans} as the main evaluation metric as well as the feedback criterion. FID is a widely-used evaluation metric for the image generation task. 
As shown in empirical studies~\cite{xu2018empirical,heusel2017gans}, FID evaluates the generation quality more effectively compared to other metrics such as Inception Score~\cite{salimans2016improved}. It is also one of the most commonly used evaluation metric in recent works. 
Our algorithm however is not tied to the specific evaluation criterion. 
That is, if there is a better evaluation metric, FID can be substituted with the new evaluation metric.

\subsection{CIFAR-10}

\noindent\textbf{Implementation Details}
We train initial candidates for 100k iterations and train each new candidate with 100k iterations after weight inheritance. We gradually increase the resolution from $d_0=8$ to 32. After reaching the final resolution, following \cite{karras2018progressive}, we further train the fixed architecture longer to achieve convergence. We follow the same training schedule as ProgGAN.

\noindent\textbf{Comparison with ProgGAN at Multiple Resolutions}
We reduce the FID from 18.33 to 12.10 at 32$\times$32, from 11.68 to 6.40 at 16$\times$16, and from 4.02 to 1.96 at 8$\times$8. The improvements compared to ProgGAN are 51\%, 45\%, and 34\% respectively. FID at each resolution is not comparable.

\noindent\textbf{Comparison with Automatic GANs}
The bottom rows in Table~\ref{tlb:cifar10} are automatic GANs~\cite{wang2019agan,gongautogan}.
Our DGGAN outperforms both and achieves the  state-of-the-art on FID.

\noindent\textbf{Comparison with State-of-the-art} 
The prior state-of-the-art methods in the top rows in Table~\ref{tlb:cifar10} are manually-designed GANs that may have improved loss functions~\cite{wang2018improving,tran2018dist} or regularization~\cite{zhang2019consistency}. These sophisticated losses and  techniques show superiority over the basic WGAN-GP loss.
Trained with only basic WGAN-GP loss, our method still outperforms all of them on FID.

\begin{figure*}[!tbp]
\begin{center}
\includegraphics[width=0.95\linewidth]{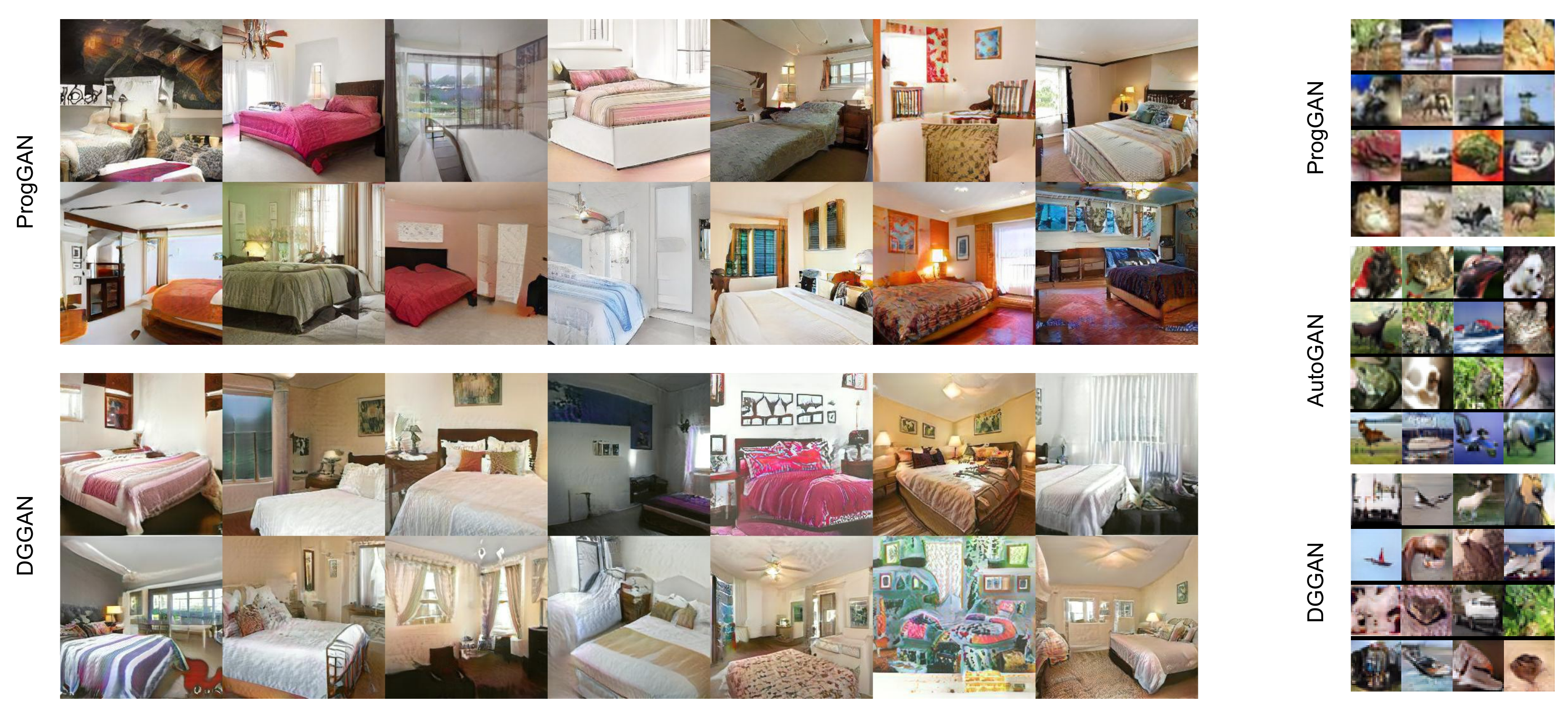}
\end{center}
  \caption{Examples of generated LSUN images (left) and CIFAR-10 images (right)\protect\footnotemark.}

\label{fig:lsun_cifar}
\end{figure*}

\noindent\textbf{Inception Score and Visualization}
Even though it is shown~\cite{xu2018empirical,heusel2017gans} that Inception Score fails to capture the distance between generated images and real images,
we still evaluate with Inception Score for a thorough comparison with prior works on CIFAR-10. 
We show that with the large variety on FID, the Inception Scores are similar across well-performing GANs.
We show randomly selected generated examples from our DGGAN, ProgGAN, and AutoGAN as in Fig.~\ref{fig:lsun_cifar}(right).

\begin{table*}[!tb]
\begin{center}
\begin{tabular}{lll}
\toprule
Model & FID & Inception Score \\
\midrule
CR-GAN~\cite{zhang2019consistency}& 14.56 & - \\
Improving MMD-GAN~\cite{wang2018improving}  & 16.21  & 8.29 \\
DistGAN~\cite{tran2018dist}  & 17.61 & -  \\
ProgGAN~\cite{karras2018progressive}  & 18.33\footnotemark & 8.80 $\pm$ 0.05  \\
WGAN-GP~\cite{gulrajani2017improved} & 29.3 & 7.86 $\pm$ 0.07  \\
DCGAN~\cite{radford2015unsupervised} & 37.7  & 6.64 $\pm$ 0.14  \\
\midrule
AutoGAN~\cite{gongautogan} & 12.42 & 8.55 $\pm$ 0.10  \\
AGAN~\cite{wang2019agan}& 23.80  &8.82 $\pm$ 0.09  \\
Ours & 12.10 & 8.64 $\pm$ 0.06  \\
\bottomrule
\end{tabular}
\end{center}
\caption{Quantitative evaluation on CIFAR-10. Top: manually-designed GANs. Bottom: automatic GANs. Lower FID and higher Inception Score indicate better generation quality.}
\label{tlb:cifar10}
\end{table*}

\subsection{LSUN}

We follow the same strategy as in CIFAR-10 except higher resolutions. We also report Sliced Wasserstein Distance (SWD)~\cite{karras2018progressive} to compare with some prior works.

\begin{table*}[!tb]
\begin{center}
\begin{tabularx}{0.9\linewidth}{l*{6}{X}}
\toprule
Resolution & 8$\times$8 & 16$\times$16 & 32$\times$32 & 64$\times$64 & 128$\times$128 & 256$\times$256\\
\midrule
ProgGAN & 29.07 & 23.75 & 27.9 & 19.67 & 29.35 & 10.76 \\
Ours & 24.29 & 21.49 & 19.01 & 8.25 & 13.29 & 8.22\\
Improvement   & 16\% & 10\% & 32\% & 58\% & 46\% & 24\% \\
\bottomrule
\end{tabularx}
\end{center}
\caption{Quantitative evaluation on LSUN, compared with ProgGAN at each scale.}
\label{tlb:lsun_scale}
\end{table*}

\begin{table*}[!tbp]
\begin{center}
\begin{tabular}{lcccccccc}
\toprule
\multirow{2}{*}{Model} & \multirow{2}{*}{Resolution} & \multirow{2}{*}{FID} & \multicolumn{6}{c}{SWD $\times 10^3$}\\
& & & 256 & 128 & 64 & 32 & 16 & Avg\\
\midrule
DCGAN~\cite{radford2015unsupervised}   & 64$\times$64 & 70.4 & - &  - & - & - & - & - \\
WGAN-GP~\cite{gulrajani2017improved} & 64$\times$64 & 20.5  & - &  - & - & - & - & -\\
Imp. MMD-GAN~\cite{wang2018improving} & 64$\times$64 & 12.52  & - &  - & - & - & - & - \\
TTUR~\cite{heusel2017gans} & 64$\times$64 & 9.5 & -  &  - & - & - & - & - \\
Ours & 64$\times$64 & \textbf{8.25} &  - & - & - & - & - & - \\
\midrule
ProgGAN (reported in~\cite{karras2018progressive}) & 256$\times$256 & 8.34 & 2.72 & 2.45 & 2.34 & 2.90 & 9.08 & 3.90  \\
ProgGAN (Pytorch) & 256$\times$256 & 10.76 & 3.74 & 3.78 & 3.53 & 4.04 & 1.58 & 3.33  \\
Ours & 256$\times$256 & \textbf{8.22} & 3.01 & 2.15 & 3.03 & 4.56& 1.40& \textbf{2.83} \\
\midrule
COCO-GAN~\cite{lin2019coco} (Part-based) & 256$\times$256 & \textbf{5.99} & - &  - & - & - & - & -\\
\bottomrule
\end{tabular}
\end{center}
\caption{Quantitative evaluation on LSUN. SWD averages over multiple feature scales from 256 to 16. Lower FID and smaller SWD indicate better generation quality.  COCO-GAN is a part-based GAN but all others are on part-based GANs that learns to generate the whole images.}
\label{tlb:lsun}
\end{table*}

\noindent\textbf{Comparison with ProgGAN at Multiple Resolutions}
We show significant improvement against ProgGAN baseline at each resolution---by 16\%, 10\%, 32\%, 58\%, 46\%, 24\% respectively, in Table~\ref{tlb:lsun_scale}.

\noindent\textbf{Comparison with Automatic GANs}
We are the first automatic GAN paper demonstrated with this high-resolution of $256\times256$. Prior automatic GAN work uses at most $48\times48$ images, potentially because of the instability of GANs, as discussed in~\cite{gongautogan}. We however train the architecture found by AutoGAN for $32\times32$ for a comparison. It obtains a FID of 19.74 on $32\times32$ and 19.61 on $64\times64$. For higher resolutions, the training fails with large patches of artifacts.

\noindent\textbf{Comparison with State-of-the-art}
Recent works~\cite{wang2018improving,heusel2017gans} on manually-designed GANs innovate on loss function and training strategy. 
These methods show superiority over the basic WGAN-GP loss.
With the disadvantage of loss, our method achieves the best FID on both resolutions among these methods as in Table~\ref{tlb:lsun}.
Different from the above models that generates the full images, COCO-GAN~\cite{lin2019coco} is a part-based method. It is trained with image parts instead of the full image, conditioned on the spatial coordinates. 
It achieves the state-of-the-art, better than other non-part based methods including ours. Combining with part-based method can be our future work with dynamic growing. 

\addtocounter{footnote}{-2}
\stepcounter{footnote}\footnotetext{\label{foot:2}All numbers are from original papers except that this FID is obtained by the PyTorch implementation}

\stepcounter{footnote}\footnotetext{\label{foot:3}ProgGAN on LSUN and AutoGAN on CIFAR-10 are obtained from their papers.}

\noindent\textbf{Visualization}
We show examples of generated bedroom images in Fig.~\ref{fig:lsun_cifar}(left). Our example images are randomly generated. We see that our generated images are diverse, sharp and have good layouts most of the time. Compared to ProgGAN, it also shows better details such as wall decorations and furniture other than beds. However, we also see failure cases such as non-straight lines and occasionally completely-failed images.

\section{Analysis}

We conduct a thorough analysis of thousands of models produced by our search. Several practical conclusions and discussions are provided. More analysis can be found in the supplementary material.

\begin{figure*}[!tb]
\begin{center}
\includegraphics[width=1\linewidth]{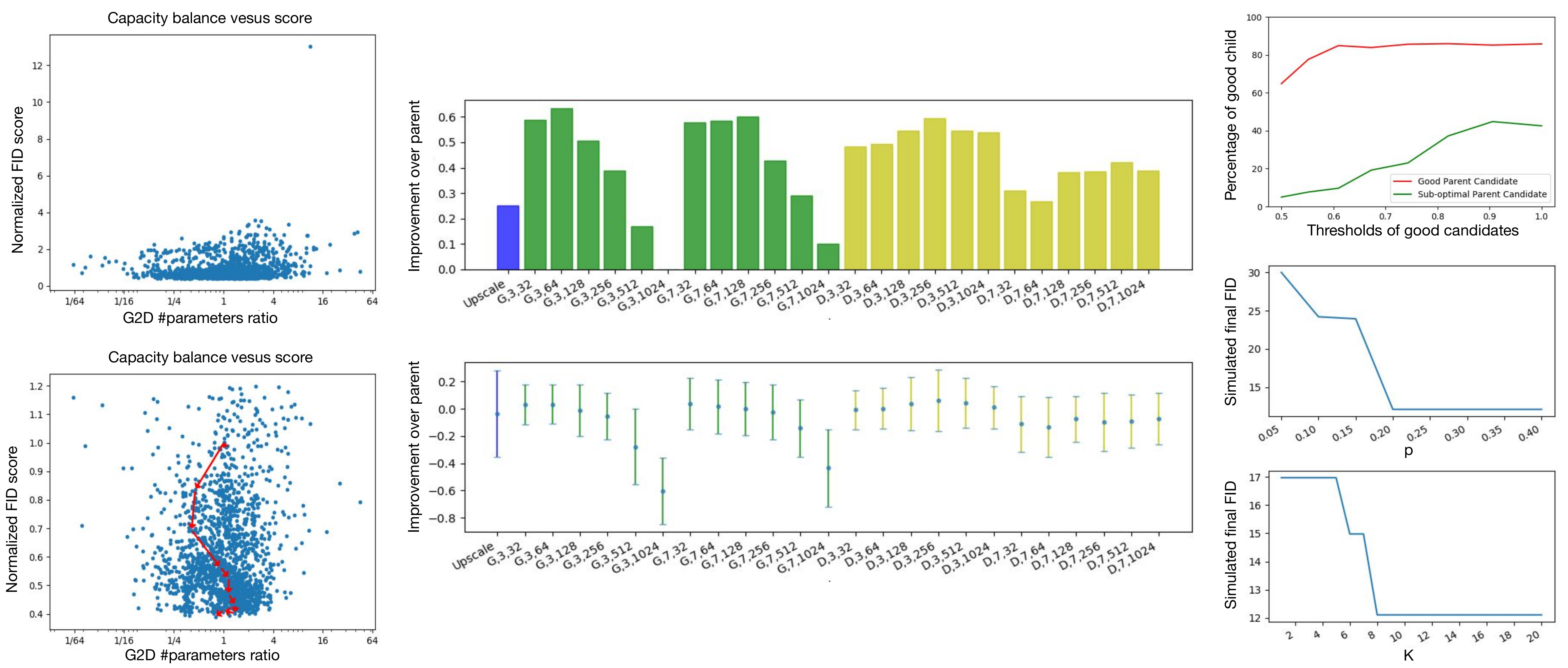}
\end{center}
  \caption{Left: G-to-D number of parameters ratio and normalized FID for candidates on CIFAR-10. The top figure includes all the candidates and the bottom figure includes the candidates with normalized FID smaller than 1.2. The red arrows show the growing process of the best performing GAN.
  Middle: How each action affects training.
  The top figure shows the percentage of candidates with positive improvement over parent after each action. The bottom figure shows the average and standard deviation of the improvement.
  Right: Ablated simulations. The top figure studies how likely good/sub-optimal parent candidates result in good child candidates with different normalized FID thresholds.  The bottom two are simulation with hyper-parameters $p$ and $K$.
  }
\label{fig:analysis}
\end{figure*}

\noindent\textbf{How does the capacity of the generator (G) and the discriminator (D) affect training?}
We use the number of parameters in G  over the number of parameters in D, noted as G2D \#parameters ratio, as the G-D capacity indicator and use FID normalized by the per-scale ProgGAN baseline as the generation quality indicator. 
We show how the candidates perform with the G-D ratio in Fig.~\ref{fig:analysis}(left). The extremely large normalized FIDs are the cases where the training diverges, which are rare with our algorithm.
From the zoomed-in figure in Fig.~\ref{fig:analysis}(left), we observe that \emph{G and D do not have to be symmetric to gain good performance}.

To have a better performance than the baseline, i.e. normalized FID$<$1, the G2D ratio can be in a wide range $[1/64, 64]$. This observation suggests that the long-existing strategy of designing symmetric GANs may miss many potentially good architecture candidates. Our DGGAN fills in this gap and explores those candidates.
To further analyze the growing process, we visualize the best performing GAN's growing route as red arrows in the plot.

\noindent\textbf{How does each action affect training?}
We investigate how each action, i.e. how to grow layers, affects training.
We compute the improvement of normalized FID over parent architectures after each action to explore how each action affects training, as in Fig.~\ref{fig:analysis}(middle). The normalized FID improvement is calculated by normalized FID of a child candidate over normalized FID of its parent candidate. 
We compute the percentage of candidates with positive improvement after each action, and the mean and variance of the normalized FID improvement of all child candidates generated by performing an action as shown in Fig.~\ref{fig:analysis}(middle).

We observe several interesting tendencies. In discriminators, larger filter size performs much worse than smaller filter size while different filter sizes perform similarly in generators. It suggests that \emph{when designing a discriminator, a small filter size should be adopted; when designing a generator, different filter sizes are worth to explore}.
We also observe that in both discriminator and generator, more filters do not always lead to better performance. \emph{Generator is especially more sensitive to over-complicated convolution layers}.

\noindent\textbf{How does greedy pruning affect searching?}
Our greedy pruning method prunes most of the sub-optimal parent candidates and only expand child candidates from the top-performing parent candidates. It accelerates the search significantly, from exponential to linear. 
However, this greedy pruning method bares the risk to miss a good search path:
a sub-optimal parent candidate that is pruned may be able to develop well-performing children models in the future.  

To quantify such risks, we show how likely the sub-optimal parent candidates could generate good children. 
We compute the ratio of good children over all children for each good or sub-optimal parent candidate using 2K+ models we trained during an extensive search process. 
The good candidates given a threshold are defined as candidates with normalized FID to be above the threshold and the sub-optimal ones are below the threshold. Each threshold holds for both parent candidates and child candidates. Note that the sub-optimal parent candidates used here are not too bad because the worst candidates have been pruned out during the search. 

We vary the normalized FID threshold in the range $[0.5,1]$ and show the results in Fig.~\ref{fig:analysis}(right).
It shows that good parent candidates are much more likely to have good child candidates, regardless of the threshold.
This indicates that \emph{pruning the sub-optimal parent candidates does not introduce much risk of missing good growing routes in practice}.

\noindent\textbf{Discussion on Efficiency}
\label{sec:efficiency}
We use hyper-parameters $K$ (top-K) and $p$ (random sampling ratio) to budget the computation cost. 
Larger $K$ or $p$ leads to a better exploration of the search space but also greater cost. 

In our experiments, we choose $K=20$ and $p=0.4$ to maximally use our computation budget to explore a larger space and more candidates to analyze the behavior of GAN's dynamic growing. 
The resulting computational cost is 580 GPU days for 2k+ CIFAR-10 models and 1720 GPU days for 1k+ LSUN models.

For the optimal time cost to achieve good performance, $K$ can be as small as 8 to achieve the same performance, and $p$ can be as small as 0.2, as shown in our simulation in Fig.~\ref{fig:analysis}(right). 
We simulate to use smaller $K$ or $p$ in our algorithm: if a simulated candidate is not among the candidates that we have reached during our real search, we ignore it. This simulation explores fewer candidates than a real run with different $K$ and $p$, which means that the actual computation cost can be less to achieve the same performance.

Note that our main contribution is to propose the dynamic growing method that bridges the gap between high-resolution GAN and architecture search. We aim for high generation quality instead of good search efficiency. Upon our simple strategy such as random sampling and pruning, further work such as using learning-based methods~\cite{liu2018darts,pham2018efficient} can be used to improve efficiency.

\section{Conclusion and Limitations}
\label{sec:conclusion}
We propose DGGAN, growing the network architecture and optimizing its parameters together automatically.
Experimental results on two datasets demonstrate competitive performance in image generation. In addition, we are the first automatic GAN method that works with high resolution images.
With a thorough analysis, we provide several constructive insights on GAN architecture designs.
We also observe several limitations as well as future directions: utilizing the newest loss functions and regularization methods rather than WGAN-GP; better evaluation criterion; more efficient search algorithms.

\bibliography{dggan}

\end{document}